\documentclass[10pt,letterpaper]{article}
\usepackage[top=0.85in,left=2.75in,footskip=0.75in]{geometry}

\usepackage{amsmath,amssymb}

\usepackage{changepage}

\usepackage[utf8x]{inputenc}

\usepackage{textcomp,marvosym}

\usepackage{cite}

\usepackage{nameref,hyperref}

\usepackage[right]{lineno}

\usepackage{float}

\usepackage{microtype}
\usepackage{booktabs}
\usepackage{tabularx}
\usepackage{caption}
\usepackage{url} 
\usepackage{subcaption}
 \usepackage{array,multirow}
\usepackage{tabularx,ragged2e,booktabs,caption,array,multirow,multicol}
\usepackage{easyReview}
\usepackage{tcolorbox}
\usepackage{amsmath}
\usepackage{nccmath}
\usepackage{makecell}
\usepackage{setspace}
\usepackage{enumitem}

\usepackage{rotating} 

\usepackage{microtype}
\DisableLigatures[f]{encoding = *, family = * }

\usepackage{array}

\newcolumntype{+}{!{\vrule width 2pt}}

\newlength\savedwidth

\raggedright
\usepackage[aboveskip=1pt,labelfont=bf,labelsep=period,justification=raggedright,singlelinecheck=off]{caption}

\makeatletter
\renewcommand{\@biblabel}[1]{\quad#1.}
\makeatother

\usepackage{lastpage,fancyhdr,graphicx}
\usepackage{epstopdf}
\fancyheadoffset[L]{2.25in}
\fancyfootoffset[L]{2.25in}
\begin{document}
\vspace*{0.2in}

\begin{flushleft}
{\Large
\textbf\newline{Deep learning for COVID-19   topic modelling via Twitter: Alpha, Delta and Omicron} 
}
\newline
\\
Janhavi Lande\textsuperscript{1},
Arti Pillay\textsuperscript{2},
Rohitash Chandra \textsuperscript{3},  
\\
\bigskip
\textbf{1} Department of Physics, Indian Institute of Technology Guwahati, Assam, India
\\
\textbf{2} School of Sciences, Fiji National University, Suva, Fiji
\\
\textbf{3} Transitional Artificial Intelligence Research Group, School of Mathematics and Statistics, UNSW Sydney, NSW 2052, Australia
\\
\bigskip

%
%



* Corresponding author: arti.pillay@fnu.ac.fj\\
* Corresponding author: rohitash.chandra@unsw.edu.au

\end{flushleft}
\section*{Abstract}

Topic modelling with innovative deep learning methods has gained interest for a wide range of applications that includes COVID-19. Topic modelling can provide, psychological, social and cultural insights for understanding human behaviour in extreme events such as the COVID-19 pandemic. In this paper, we use prominent deep learning-based language models for COVID-19 topic modelling taking into account data from  emergence (Alpha) to the Omicron variant. We apply topic modeling to review the public behaviour  across the first, second and third waves based on Twitter dataset  from India.  Our results show that the topics extracted for the subsequent waves had certain overlapping themes such as covers governance, vaccination, and pandemic  management while novel issues aroused in political, social and economic situation during COVID-19 pandemic. We also found a strong correlation of the major  topics qualitatively to news media prevalent at the respective time period. Hence, our framework has the potential to capture major issues arising during different phases of the COVID-19 pandemic which can be extended to other countries and regions.



{The 2019 coronavirus pandemic (COVID-19) \cite{fauci2020covid,velavan2020covid} has received the most widespread media coverage in comparison to the other pandemics, such as severe acute respiratory syndrome (SARS) (2002 – 2003) \cite{zhong2003epidemiology}, Asian flu (1957–1958) \cite{jackson2009history}, and  the Black death (1347–1351) \cite{green2015pandemic}. COVID-19 is also known as also known as  (SARS-CoV-2) \cite{pitlik2020covid}. COVID-19 pandemic had major challenges on  mental health during 
lockdowns \cite{doi:10.1177/0020764020950769}, economic consequences due to closure of businesses and  job losses, distribution of rations \cite{rations.2021}, and  administration of vaccines, etc \cite{Kumar2021}. The World Health Organisation (WHO) first reported COVID-19 as a public health crisis in January 2020 and declared  it a pandemic in March 2020 \cite{lwin2022evolution}. India reported its first case on January 3rd, 2020 with a increase in cases reported from March also declared as India’s first wave \cite{sarkar2021covid}. India also reported its first COVID-19 death in March, 2020 \cite{kumar2021covid}. During the first wave in India several measures were taken by Ministry of Health and family welfare and Government of India in collaboration with WHO to prevent and control spread of the virus within and across states of India. Some measures included personal protection measures, closures of educational institutions, social distancing measures, closures of historic buildings, transport measures to stop and control flights, trains and other domestic transportation and measures to control sports activities \cite{kumar2021covid}. Despite the implementation of these measures, several challenges were faced in controlling the spread of virus and management of treatment of patients. The  challenges  included overburdened and understaffed hospitals, lack of medical equipment, fake messages about the pandemic circulated in media, frustrations from public related to lockdowns, cancellation of annual prayer and sports events and controlling the rush, and over crowding in supermarkets to stock up rations \cite{kumar2021covid}. However, these were not unique to India in comparison to other COVID-19 struck countries. In the case of India, the return of migrant workers to their respective states also reported an increase in COVID-19 cases in the respective states \cite{kumar2021covid} . During this first wave, Government of India also announced  20 \textit{lakh crore Rupees} (270 billion USD)  package \cite{Indiacovidreliefpkg} of which 1.7 lakh crore Rupees  was used for food rations  for those most severely affected by the lockdown. In addition, approximately 800 million also benefited from free grain, cash, and cooking gas benefits for three months which addressed the fear of food security in India by WHO \cite{mishra2020covid}.

Social media became one of the major platforms to spread information on COVID-19 \cite{goel2020social}; with activities such as disseminating information on prevention, control and screening centers. Social media has also been used for reporting death rates, and providing advice on lockdowns etc. Additionally it became a public space to share and capture positive, negative and neutral sentiments related to the management of the pandemic by the respective agencies \cite{gonzalez2020social,venegas2020positive}.  Twitter, a micro-blogging social media platform, was widely used to share and express personal viewpoints and  sentiments during the various waves of COVID-19 \cite{chandra2021covid}. Due to publicly available data (tweets) provided by Twitter, it has been used for social media analysis during various events such as elections \cite{chandra2021biden}, protests, natural disasters, endemics, and also the COVID-19 pandemic \cite{chandra2021covid}.}

 
Topic modelling is an area in natural language processing (NLP)  that learns the topics of a documents, analyses and categories them according to common patterns and themes emerging across the documents \cite{tong2016text,nikolenko2017topic}. Other than information retrieval and analysis, topic modelling also allows researcher to identify influential persons and groups on a specific social media platform. It can also be used to detect any signs of adverse mental issues such as depression \cite{jonsson2015evaluation}. It has been widely used as a quantitative and qualitative research and analysis tool in the field of bioinformatics \cite{liu2016overview}, management \cite{storopoli2019topic}, sociology, opinion and media analysis \cite{nikolenko2017topic}. It also enables researchers to carrying out a smart literature review by categorically compiling literature while avoiding the onerous task of a manual review \cite{asmussen2019smart}. The two traditions techniques for topic modelling include Latent Semantic Analysis (LSA) and Latent Dirichlet Allocation (LDA) \cite{jonsson2015evaluation,asmussen2019smart}.  LDA \cite{blei2003latent} assumes that documents are a mixture of topics and each topic is a mixture of words with certain probability score. Though these techniques have been widely used over the years for data extraction and analysis, it is not highly suitable for short text documents such as Twitter \cite{jonsson2015evaluation}. Twitter allows users to send and receive 140-character short messages (tweets) making it one of the fastest and popular data acquisition sources \cite{tong2016text}.
Researchers have used deep learning models such as Long Short-term memory (LSTM) networks   \cite{hochreiter1997long}  and bi-directional encoder representative from transformer (BERT) \cite{devlin2018bert}  models that are increasingly becoming popular for language modelling tasks that include sentiment analysis and topic modelling. In a study, Chandra and Krishna \cite{chandra2021covid}  implemented LSTM and BERT models for  sentiment analysis of Twitter data for the rise of COVID-19 cases in India and reported that the optimistic, annoyed and joking tweets mostly dominated the monthly tweets with much lower portion of negative sentiments. In a recent study,  Chandra and Ranjan \cite{chandra2022artificial} used a BERT-based model with advanced clustering methods for topic modelling by comparing topics of ancient Hindu texts, i.e the Bhagavad Gita and the Upanishads. In this study, there were different combinations of BERT-based models with clustering methods, which provided better results when compared to LDA. Hence, this motivates us to use the same framework to compare topics emerging across the three major peaks for COVID-19 in India, which had unique set of challenges.

 In this paper, we use deep learning based language framework for COVID-19 topic modelling taking into account data from COVID-19 emergence, which includes Alpha, Delta and Omicron variants for the three distinct peaks (waves) in India as a case study. We use Twitter data from India and also compare with our earlier works that looked at sentiment analysis of the first wave in India. We note that we refer to the three distant peaks as waves, we define the timeline in the methodology section. Our goal is to extract and study the various topics emerging in the three different waves and discuss the relationship to emerging events and issues in the media during the respective time-frames. Our goal is to see how these affected the topics covered in COVID-19 tweets during the respective timelines. 
  
The rest of the paper is organised as follows. In Section 1, we provide further details about the methodology which includes data processing and models. In Section 2, we present the results and Section 3 provides a discussion about the results in relation to emerging events in the respective waves. Section 4 concludes  the paper with directions for subsequent research.

\section{Methodology} 

\subsection{COVID-19 Waves in India}

 Whilst India had a prolonged first wave which lasted almost a year, it recorded lower COVID-19 prevalence (in terms of number of cases when compared to other countries, i.e USA and UK \cite{sarkar2021covid}. India had much lower number of cases and deaths per capita when compared to developed western countries \cite{kundu2020covid}; i.e as of July 12th 2020, India verified 820,916 COVID-19 cases, and 22,123 deaths and the mortality rate of India was 2.69\%. In comparison, US had 3,097,300 confirmed cases and 132,683 deaths with a mortality rate of 4.28\% and the UK had  288,137 cases and 44,650 deaths with mortality rate of 15.49\%. Furthermore, India recorded its second wave with a major and much higher peak of cases that spanned March to July 2021 \cite{sarkar2021covid}. In the first wave, the peak was reached with around 97,000 novel daily cases (16th September, 2020) \cite{dong2020interactive}. In comparison to the first wave, the second wave spread rather rapidly with a peak of more than 400,000 cases reported per day (7th May 2021) from around 9,000 cases (15th February, 2021) within three months \cite{dong2020interactive}. The Delta variants, i.e., B.1.617.1 and B.1.617.2 \cite{mlcochova2021sars} also emerged during this period being one of the major reasons for the rapid increase in cases. This caused serious concerns in the international community and call for addressing medical support and food insecurity \cite{kuppalli2021india}. Afterwards, there was a steady decline in infections mostly attributed to better control and management of the virus and the administration of vaccines \cite{kumar2021strategy}. The third wave recorded from late December 2021 with a major peak of about 306,000 novel daily cases (23rd January, 2022), given the Omicron variant \cite{he2021sars}. The definition of major variants has  been given in \cite{covidclassification}
 




 
 

 \subsection{ Data Extraction and Processing}

 We obtained the dataset of tweets originating from India during COVID-19 first-wave from the IEEE Dataport \cite{781w-ef42-20} for India.

 We note that Twitter does not allow the tweets to be shared directly with third parties; hence many associated dataset generally feature the tweet identifiers (IDs) and there are tools (known as hydrator    \url{https://github.com/DocNow/hydrator}.) that can be used to extract the tweets which is a time consuming process due to restrictions given by Twitter to ensure that the data is not misused. We obtained the dataset of tweet handles (i.e tweet identifier) for the second and third waves from 'Coronavirus (COVID-19) Tweets Dataset' from IEEE Dataport, which features  more than 310 million COVID-19 specific English language tweets. We also published the dataset via Kaggle that we obtained which features major countries and the tweets from emergence to Omicron variant (till February 2022)  \cite{janhavi2022}.
 
We  used the hydrator software application to extract the daily tweets worldwide and then separated the India-specific tweets  from the global dataset. We have obtained 30,000 tweets per day from India for three selected days a week for the time-frame of second and third wave defined in Table \ref{tab:one}.  We obtained hydrated tweets with 'coordinates', 'created\_at' , 'hashtags', 'media', 'urls', 'favourite\_count','id', 'in\_reply\_to\_screen\_name', 'in\_reply\_to\_status', 'lang', 'user\_location', 'text', 'place', 'retweet\_count', 'source', 'tweet\_url', 'user\_description', 'user\_favourite\_counts' etc. We   pre-processed  the entire dataset consisting of the three waves, which consists of the following steps.
\begin{itemize}[noitemsep]
\item Removal of punctuation and web-links (URLs); 
\item Removal of mentions in tweets;
\item Conversion of symbols used for emotions (emojis) into text using emoji2text \footnote{\url{https://pypi.org/project/emoji2text/}}, ;
\item Removal of extra spaces, and lower-casing and removal of stopwords;
\item Lemmatisation process to group similar words using the    natural language toolkit (NLTK) library.  \footnote{\url{https://www.nltk.org/}};

\end{itemize} 
Lemmatization, removal of stop words and stemming is not required with Top2Vec model \cite{angelov2020top2vec}.

\begin{table*}
\small
\centering
\begin{tabular}{||c|c|c|c|c|c|c|c||}
\hline
Corpus & \# Tweets & \# Words & \#Cases & \#Major Variants & \#Mutants \\ \hline
\hline
First wave & 159312 & 2468018 & 10302012 & Alpha & E484Q and E484K\\
\hline
Second Wave & 187531 & 12274701 & 24207004 & Delta, Delta Plus   & K417N\\
\hline
Omicron & 172583 & 4589488 & 7652375 & Omicron & -- \\

\hline
\end{tabular}
\caption{Dataset Statistics}
\label{tab:one}
\end{table*}
     
\subsection{Models}
\subsubsection{LDA}
LDA is a prominent generative model for discovering the abstract  topics  that occur in a collection of documents. Hence, LDA has been prominent for topic modelling \cite{jelodar2019latent} and been applied to studies such as bioinformatics  \cite{liu2016overview}, social media user recommendation  \cite{pennacchiotti2011investigating}, and scientific paper recommendation \cite{amami2016lda}. LDA  builds a topic per document model and words per topic model, modeled as Dirichlet distributions \cite{blei2001latent,blei2003latent}. LDA allows sets of observations to be explained by unobserved groups; for example, if observations are words collected into documents, it posts that each document is a mixture of a small number of topics. LDA had challenges when it comes to social media such as Twitter due to short size of the tweets; however there has been some success \cite{negara2019topic} and certain amendments have also worked \cite{resnik2015beyond}.

In LDA, $\alpha$ and $\eta$ are proportion parameter and topic parameter, respectively. The topics are given by  $\beta_{1: K}$, where each $\beta_{k}$ is a distribution over the vocabulary. The topic proportion for the $d$ the document are $\theta_{d}$ , where $\theta_{d, k}$ is the topic proportion for topic $k$ in document $d$. The topic assignments for the $d$ th document are $Z_{d}$, where $Z_{d, n}$ is the topic assignment for the $n$th word in document $d$. Finally, the observed words for document $d$ are $w_{d}$, where $w_{d, n}$ is the $n$th word in document $d$, which is an element from the fixed vocabulary.



The topic assignment $Z_{d, n}$ depends on the per-document topic distribution $\theta_{d}$; and the word $w_{d, n}$ depends on all of the topics $\beta_{1: K}$ and the topic assignment $Z_{d, n}$.

 \subsubsection{GSDMM}
 
 The Gibbs sampling Dirichlet multinomial mixture model (GSDMM) \cite{yin2014dirichlet} works well with short text clustering which is a major source of data given social media \cite{rangrej2011comparative}. Short text clustering is challenging since usually a single tweet consists of a single topic of unigrams. Mazarura and Waal  \cite{mazarura2016comparison} presented a comparison of the performance of LDA and GSDMM on short text and reported  that the LDA generally outperformed GSDMM on the long text and on the short text, GSDMM displayed better potential. The model claims to solve the sparsity problem of short text clustering while also displaying word topics like LDA. GSDMM is essentially a modified LDA (Latent Dirichlet Allocation) which assumes that a document (tweet or text for instance) encompasses one topic. This differs from LDA which assumes that a document can have multiple topics. 
 Hu et. al. \cite{HU2022239} showed that GSDMM  has better performance than related methods for Web service clustering. The generative process for GSDMM can be expanded for the whole corpus as follows:
 \begin{enumerate}[noitemsep]
\item Randomly choose $T$ topic distributions, $\vec{\beta}_{t} \sim$ $\operatorname{Dirichlet}\left(\vec{\lambda}_{\beta}\right)$.
\item Randomly choose a distribution over topics, $\vec{\alpha} \sim$ Dirichlet $\left(\vec{\lambda}_{\alpha}\right)$, for the corpus.
\item For each document, $\vec{d}$ where $d=1,2, \ldots, D$ :
a) Randomly choose a topic $z_{d} \sim \operatorname{Multinomial}(\vec{\alpha})$.
b) Randomly choose words $w_{d} \sim \operatorname{Multinomial}\left(\vec{\beta}_{z_{d}}\right)$.
 \end{enumerate}
The collapsed Gibbs sampler was developed based on the following rationale. Whilst conditional distributions can be derived for all the variables, since $\vec{z}=\left\{z_{1}, z_{2}, \ldots, z_{T}\right\}$ is a sufficient statistic for $\vec{\alpha}=\left\{\alpha_{t}\right\}_{t=1}^{T}$ and $\beta=\left\{\vec{\beta}_{t}\right\}_{t=1}^{T}$, they can both be calculated from $\vec{z}$. Consequently, if the parameters $\vec{\alpha}$ and $\beta$ are integrated out of the posterior distribution, $p\left(\vec{\alpha}, \beta, \vec{z} \mid \lambda_{\alpha}, \lambda_{\beta}\right)$, we simply need to sample from $\vec{z}$. 

Ultimately, the estimation of the parameters $\vec{\alpha}=\left\{\alpha_{t}\right\}_{t=1}^{T}$ and $\beta=\left\{\beta_{w t}\right\}_{t=1, w=1}^{T, V}$ is simplified to calculating $\beta_{w t}=$ $\frac{n_{t}^{w}+\lambda_{\beta}}{\sum_{w=1}^{V} n_{t}^{w}+V \lambda_{\beta}}$ and $\alpha_{t}=\frac{m_{t}+\lambda_{\alpha}}{\sum_{t=1}^{T} m_{t}+T \lambda_{\alpha}}$ where $V$ is the size $\sum_{\text {of the vocabulary, } n_{t}^{w}}^{w}$ is the number of occurrences of word $w$ in topic $t, m_{t}$ is the number of documents in topic $t$ and $\vec{n}_{t}=\left\{n_{t}^{w}\right\}_{w=1}^{V}$.

 \subsubsection{BERT for Topic Modelling via Clustering}
 
 BERT is pre-trained language model \cite{devlin2018bert}  which is based on Transformers, i.e. encoder-decoder LSTM-based recurrent neural network that features enhanced memory mechanism known as attention \cite{vaswani2017attention}.  The encoder  generates an encoding that feature information about the relevant parts of the inputs, which is passed to the next encoder layer as inputs. The  decoder layer does the opposite of the encoder to generate an output sequence, and each encoder and decoder layer makes use of attention mechanism. 
Transformer models implement the mechanism of attention by  weighting the significance of each part of the input data, which has made them prominent for language modelling tasks \cite{beltagy2020longformer}.

 BERT pre-training phase involves semi-supervised learning  tasks such as masked language modelling \cite{devlin2018bert,liu2019roberta,lan2019albert} and next sentence prediction \cite{devlin2018bert}. The two BERT variants include  $BERT_{BASE}$ which consists of 12 transformer blocks  and a total of 110 million parameters,  and  $BERT_{LARGE}$ which  consists of of 24 transformer blocks with   340 million parameters.  Although BERT is pre-trained, it is usually  trained further with datasets for specific applications, such as sentiment analysis of COVID-19 related tweets during the rise of novel cases   \cite{chandra2021covid}, and  modelling USA 2020 presidential elections \cite{chandra2021biden}, and sentiment analysis as a means to compare translations of  religious and philosophical texts \cite{chandra2022semantic}.

Clustering methods refer to  unsupervised machine learning that  groups unlabelled data based on a given similarity measure. The goal of clustering  is to assign each data-point a label or a cluster identify. A large number clustering algorithms exits in literature  \cite{xu2015comprehensive}.  \textit{Hierarchical density based spacial clustering of application with noise} (HDBSCAN) \cite{campello2013density, mcinnes2017accelerated}  defines clusters as highly dense regions separated by sparse regions with the goal of finding  high probability density regions as clusters. Clustering methods can be used for topic modelling given that a word embedding is obtained from language models. Recently, several topic modelling frameworks  used BERT for embedding in combination with clustering methods \cite{angelov2020top2vec} \cite{sia2020tired} \cite{thompson2020topic} \cite{chandra2022artificial} \cite{grootendorst10bertopic}.

\subsection{Framework} 
In topic modelling, a \textit{word}  is the basic unit of data  which refers to items from dataset (vocabulary) of size $V$  indexed by $\{1, ..., V\}$. A collection of  words is known as a  \textit{document} which can be denoted as  $\mathbf{w} = \{w_1, w_2, ..., w_N\}$ for a sequence of size $N$, where $w_i$ is the $i^{th}$ word. The  collection of $M$ documents is known as a  corpus denoted by $D = \{\mathbf{w}_1, \mathbf{w}_2, ..., \mathbf{w}_M\}$\cite{blei2003latent}.%
 
\begin{figure*}[htbp!]
\centering

\begin{adjustwidth}{-2.25in}{0in} 
\includegraphics[width=17cm]{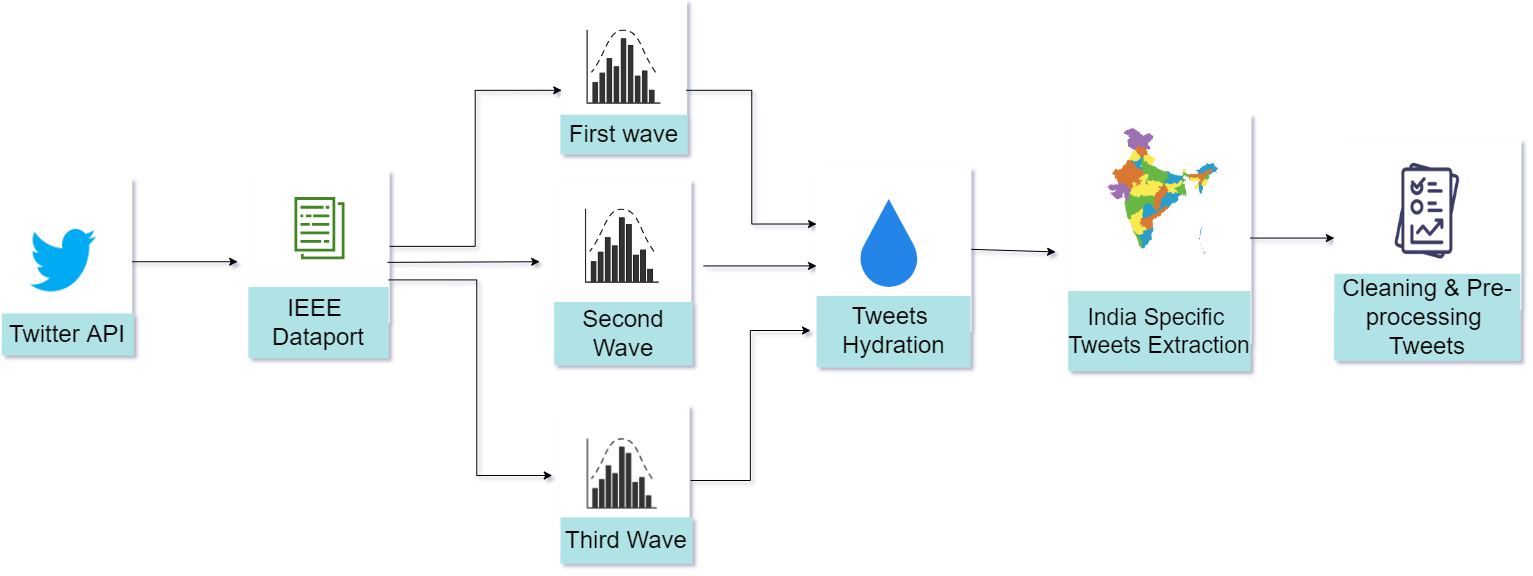}
\caption{India specific dataset extraction from global COVID-19 tweets dataset.}
\label{fig:framework1}

\end{adjustwidth}
\end{figure*}

We present a framework to employ various machine learning models for topic modeling as shown in Figure 1 and Figure 2.  Figure 1 describes the process about extraction of COVID-19 related tweets using a combination of the tweets originating from India during COVID-19  \cite{lamsal2020design} that covers the first wave in India. We obtained the tweets for the second and third waves from our global dataset \cite{janhavi2022} as described in Section 1.2. As shown in Figure  \ref{fig:framework1}, we used location based extraction to obtain India-specific tweets and pre-processed the tweets of all the three waves before  modelling. 
 
\begin{figure*}[htbp!]
\centering

\begin{adjustwidth}{-2.25in}{0in} 
\includegraphics[width=17cm]{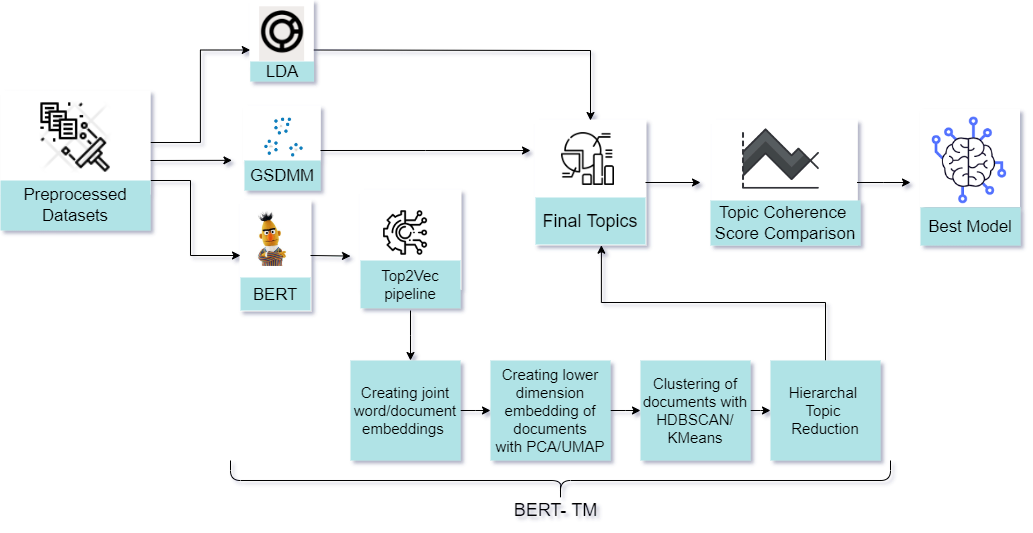}
\caption{Framework for COVID-19 topic modeling with LDA, GSDMM, BERT-TM using dataset obtained via Twitter.}
\label{fig:framework}

\end{adjustwidth}
\end{figure*}

In Figure \ref{fig:framework}, our framework begins by training the LDA, GSDMM and BERT-based topic2Vec model. Note that LDA and GSDMM can be directly used for topic modelling and hence the final topics are directly obtained. 
 The original BERT model is computationally intensive for making predictions. Sentence BERT (S-BERT)\cite{reimers2019sentence} improved 
BERT model by reducing computational time to to derive semantically meaningful sentence embedding. BERT on its own cannot be used for topic modelling, it only provides a word embedding that would be an input for clustering methods. Hence, our framework employs S-BERT via $BERT_{base}$ with    HDBSCAN and refer to it as BERT topic model (BERT-TM), hereon-wards.
We pass the word embeddings generated by S-BERT to the Top2Vec pipeline where the documents are placed close to other similar documents. Note that S-BERT is embedded within the Top2Vec model in order to obtain world embedding.  

Since our S-BERT-based word embedding model has a large vector of features, we need a dimensional reduction method to reduce the features. Hence, we use \textit{uniform manifold approximation and projection (UMAP)} \cite{mcinnes2018umap} which is a non-linear dimensionality reduction   based on Riemannian geometry and algebraic topology. UMAP can be used in a way similar to principal component analysis (PCA)\cite{wold1987principal} for visualisation and  dimensionality reduction of high dimensional data. Chandra et al. \cite{chandra2022unsupervised} evaluated prominent dimensional reduction methods with clustering for distinguishing variants of concern based on COVID-19 genome sequence and reported that UMAP as the best method for the application. Hence we use UMAP in our framework for this paper. 
 
 In Figure  \ref{fig:framework}, finally we finally obtain the  topic vectors  by taking the centroid of document vectors in the original dimension. We then perform hierarchical topic reduction of the obtained topics in order to assess similarity between the the topics of the three waves. We use gensim \footnote{\url{https://radimrehurek.com/gensim/}},a free open-source Python library for representing documents as semantic vectors, as efficiently and painlessly as possible in order to 
 obtain topic coherence scores of each model. The algorithms in Gensim are unsupervised, which means no human input is necessary – we only need a corpus of plain text documents.


\subsection{Topic Coherence}

A topic coherence measure is typically used to evaluate  topic models to measure  their ability to capture topics that can well describe the document (i.e. low perplexity) and how well the topics that carry coherent semantic meaning. The topic coherence measure  can be in two ways: 1.) rating, where human evaluators rate the topic quality on a three point topic quality score, and two.) intrusion, where each topic is represented by its top words along with an \textit{intruder} which has very low probability of belonging to that topic.\cite{https://doi.org/10.48550/arxiv.2107.02173}  
We use \textit{topic coherence}  (TC)\cite{roder2015exploring} as a metric to fine tune and evaluate different models on different corpus. The  topic coherence metric based on the \textit{normalized point wise mutual information}(NPMI) correlates really well with the human evaluation and interpretation of the topic coherence \cite{DistribSemantics}. R{\"o}der et. al.\cite{roder2015exploring} presented  a detailed study on the coherence measure and its correlation with the human topic evaluation data. 

Note that the NPMI is a step used in topic coherence measure for a pair of words $(w_i, w_j)$, from the top $N$ (set to 50) words of a given topic as given below: 
 
\begin{center}

$\vec{v}\left(W^{\prime}\right)=\left\{\sum_{w_{i} \in W^{\prime}} \operatorname{NPMI}\left(w_{i}, w_{j}\right)^{\gamma}\right\}_{j=1, \ldots,|W|}$
\\
$\operatorname{NPMI}\left(w_{i}, w_{j}\right)^{\gamma}=\left(\frac{\log \frac{P\left(w_{i}, w_{j}\right)+\epsilon}{P\left(w_{i}\right) \cdot P\left(w_{j}\right)}}{-\log \left(P\left(w_{i}, w_{j}\right)+\epsilon\right)}\right)^{\gamma}$
\\
$\phi_{S_{i}}(\vec{u}, \vec{w})=\frac{\sum_{i=1}^{|W|} u_{i} \cdot w_{i}}{\|\vec{u}\|_{2} \cdot\|\vec{w}\|_{2}}$
\end{center}
where, we compute the joint probability of the single word $P\left(w_{i}\right)$  by the Boolean sliding window approach (window length of $s$ set to the default value of 110). We create a  virtual document and count the of occurrence of the word ($w_i$) and the word pairs ($w_i, w_j$), and then it is divided by the total number of the virtual documents.


\subsection{Technical details}

In the implementation of our framework, we used pre-trained S-BERT\footnote{\url{https://huggingface.co/sentence-transformers/distiluse-base-multilingual-cased}}, which has been trained on a large corpus of 15 different languages. The model uses DistilBERT\cite{sanh2019distilbert} as the base transformer model, then its output is pooled using an average pooling layer and a fully connected (dense) layer is used finally to give a 512 dimensional output. We used different combination of dimensionality reduction techniques and clustering algorithms with the pre-trained semantic embeddings to get the final topics for each corpus.

We reduce the embedding dimension to the 5-dimension using UMAP which  uses two important parameters, \textit{n\_neighbors} and \textit{min\_dist} in order to control the local and global structure of the final projection. We  use default \textit{min\_dist} value of $0.1$, \textit{n\_neighbors} value of $10$ and the \textit{n\_components} value of $5$. We set the \textit{random-state} to $42$ and use \textit{cosine-similarity} as the distance metric.   
We later use   HDBSCAN  with  parameter $\textit{min\_samples} = 5$, which is used to estimate the probability density of the data points. The  \textit{min\_cluster\_size}   defines the smallest grouping size to be considered as cluster, we set it to $10$. Finally, in the remaining two parameters,  we use $metric=euclidean$ and $min\_samples = 5$. 


\section{Results}
\subsection{Data Analysis} 
\begin{figure*}[!htbp]
     \centering
     
\begin{adjustwidth}{-2.25in}{0in} 
     \begin{subfigure}{0.7\textwidth}
         \includegraphics[width=1.1\linewidth]{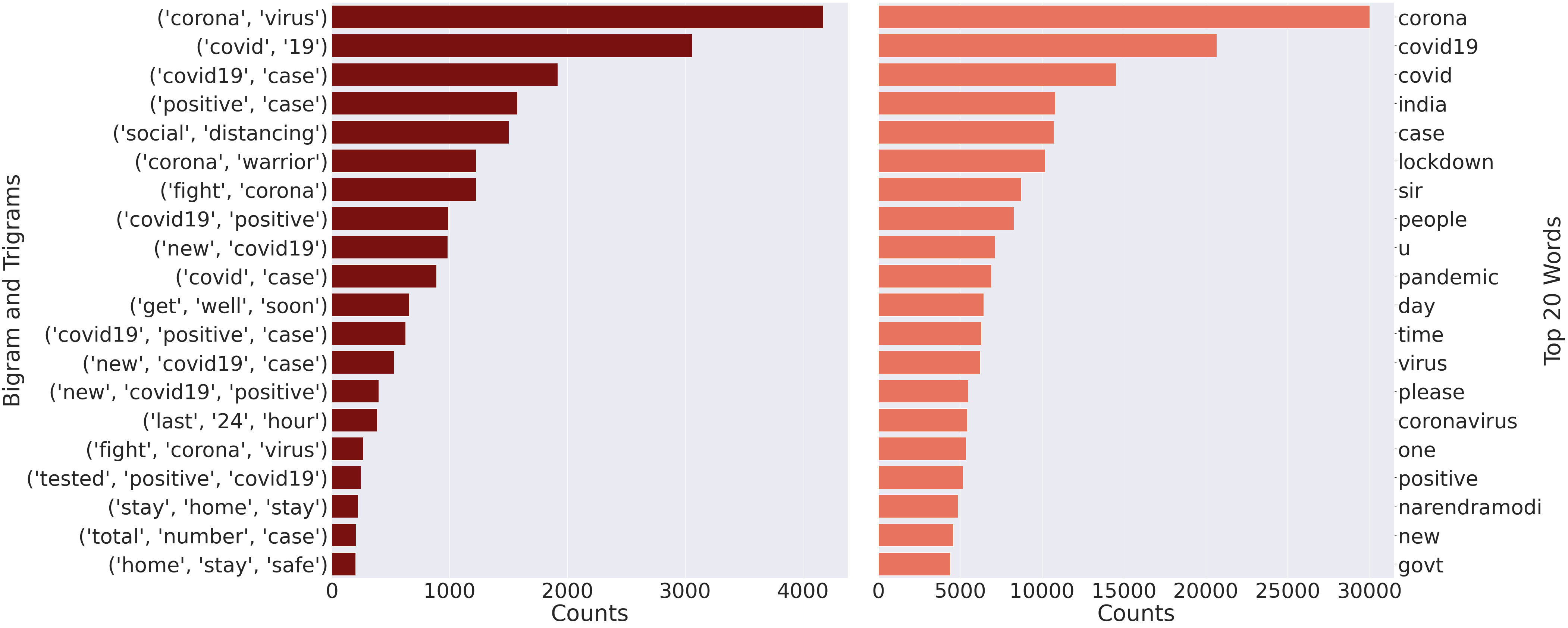}
         \caption{First wave}
         \label{fig:eknath_gita_uni_bi_tri}
     \end{subfigure}\\
     \hfill
     \begin{subfigure}{.7\textwidth}
       \includegraphics[width=1.1\linewidth]{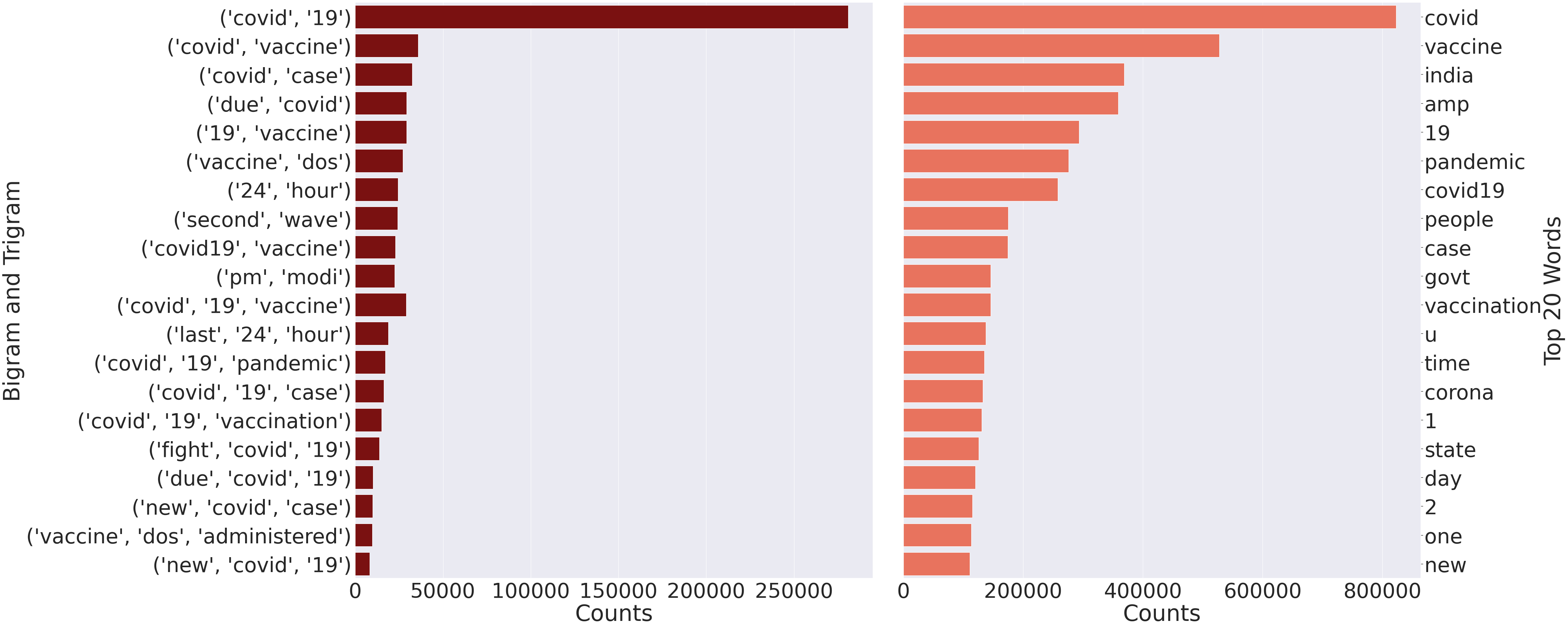}
         \caption{Second Wave}
         \label{fig:eknath_upanishads_uni_bi_tri}
     \end{subfigure}\\
      \hfill
     \begin{subfigure}{.7\textwidth}
         \centering
         \includegraphics[width=1.1\linewidth]{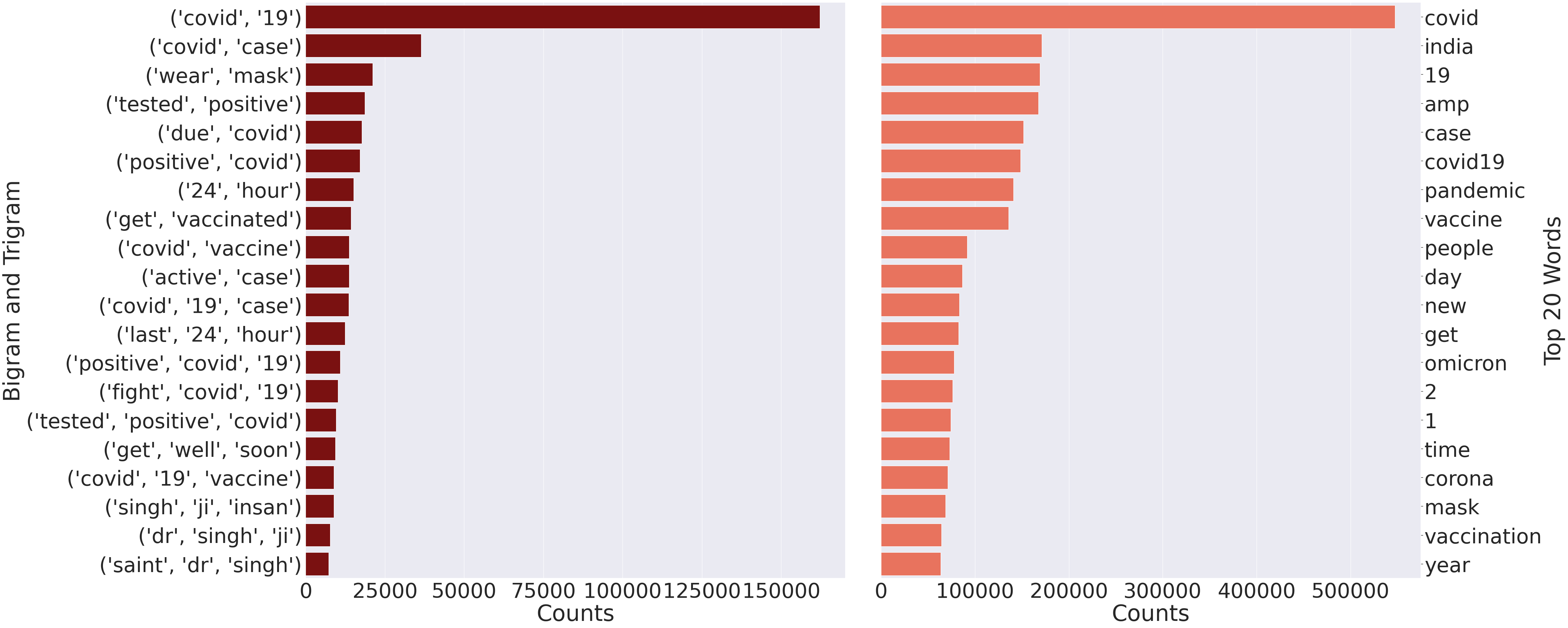}
          \caption{Third wave}
          \label{fig:ten_upanishads_uni_bi_tri}
     \end{subfigure}
     \caption{Bigrams and Trigrams   for first, second and third waves of  COVID-19 in India.}
        \label{bigramstrigrams}
        
\end{adjustwidth}
\end{figure*} 

We first present the bigram and trigram analysis of the three waves of COVID-19 in India.  Figure \ref{bigramstrigrams} - Panel (a) presents the bigrams and trigrams of the first wave. We observe that the bigrams ['corona' 'virus'] and ['covid' '19'] are more prominent which were used to refer to the pandemic followed by bigrams associated with positive case and social distancing. Hence, the tweets are centered around understanding the virus and its actual name, nature of positive cases, its preventive measures such as social distancing and staying at home. The trigrams further expand the ideas in the bigrams such as ['covid19' 'positive' 'case'], with new ones such as ['get' 'well' 'soon'] and ['last' '24' 'hour'] which were commonly expressed in media and also been part of official statements of politicians and leaders, not just in India but around the world.  Looking at the top 20 words, we find that  the word 'lockdown' is mentioned slightly fewer times, however the words mentioned more than it are merely referring  the different names of the virus, number of cases and location (India). This indicates that lockdown as a means to prevent the virus from spreading was highly discussed  during the first wave,  since the lockdowns were harshly implemented then which gained public attention in social media. 

Figure \ref{bigramstrigrams} - Panel (b) shows the bigrams and trigrams for the second wave are mostly similar to the first wave. The bigram, ['second' wave]  and trigram ['vaccine'  'dos'  'administration'] well point out the major topics that are unique for this dataset. These were discussed in the media during the second wave where the vaccine dosage administration was a major discussion topic. Apart from these, we notice ['pm' 'modi'] which refers to Prime Minister Narendra Modi who has been active in media briefings with the roll-out of vaccination programme \cite{modi.2021}. In terms of the top 20 words, the term 'vaccine' is the second most expressed which further shows how important the vaccination process has been during the second wave. In the case of the third wave in Figure \ref{bigramstrigrams} - Panel (c), the top bigrams and trigrams are similar to the first and the second waves; however, ['wear' 'mask'], ['get' 'vaccinated'], and ['tested' 'positive' 'covid'] are some of the key unique ones in the third wave. Apart from these, variations of  ['dr' 'singh' 'ji'] are also  given as trigrams, this could be pointing to Dr Poonam Khetrapal Singh who was at the time the Regional Director of the World Health Organisation for South-East Asia Region \cite{who.2021}.

\subsection{Evaluation of Topic Modelling methods }

We evaluate the respective models by topic coherence of the topics obtained. Table \ref{tab:metric} shows the NPMI as a  topic-coherence measure for different models (BERT-TM, LDA, and GSDMM) given in the framework shown in Figure \ref{fig:framework} for the three different COVID-19 waves in India which were treated as implemented datasets (documents). We trained LDA  for 200 iterations default hyper-parameters implemented in the \textit{gensim} \cite{rehurek_lrec} language modelling library. We fine-tuned the number of topic parameters to get the optimal value of NPMI. In Table \ref{tab:metric}, we observe that in the case of the first wave, BERT-TM gives the highest NPMI score indicating  better results. This is followed by GSDMM and LDA, which are closer to each other than BERT-TM. Note that the number of topics extracted by the three models are similar in range of (58 and 60).

\begin{table}[htbp!]

\begin{centering}
\small
\begin{tabular}{ccc}
\hline
\hline
Model & No. Topics & NPMI \\
\hline
\hline
BERT-TM& 58 & 0.69  \\
GSDMM & 60 & 0.39 \\
LDA& 58 & 0.34 \\ 
\hline
\hline
\end{tabular}
\caption{NPMI score (higher is better) for the first wave of the COVID-19 pandemic in India. }
\label{tab:metric}
\end{centering}
\end{table} 

Next, we use the best model (BERT-TM) obtained from Table \ref{tab:metric} using NPMI and present the topics extracted from the respective documents (waves). Note that the BERT-TM employs dimensional reduction via (UMAP) and  clustering via HDBSCAN. In comparison with k-means clustering, the major advantage of  HDBSCAN is that it   does   does not require  to specify the number of clusters that corresponds to the number of topics. BERT-TM also uses  hierarchical topic reduction to reduce the number of topics  so that they are more interpretable  \cite{angelov2020top2vec}. Apart from high topic coherence, the other advantage of BERT-TM is that it has major components that can be separately analysed, i.e. we can visualise results from dimensional reduction via UMAP, and also use other methods if needed. Furthermore, we can also visualise results from the clustering component which is implemented by HDBSCAN; hence, fighter insight in BERT-TM enables it to be  interpretable machine learning model.


\subsection{Topic modelling: First vs Second Wave}
\begin{figure}[htbp!]
\centering
\includegraphics[width=9cm]{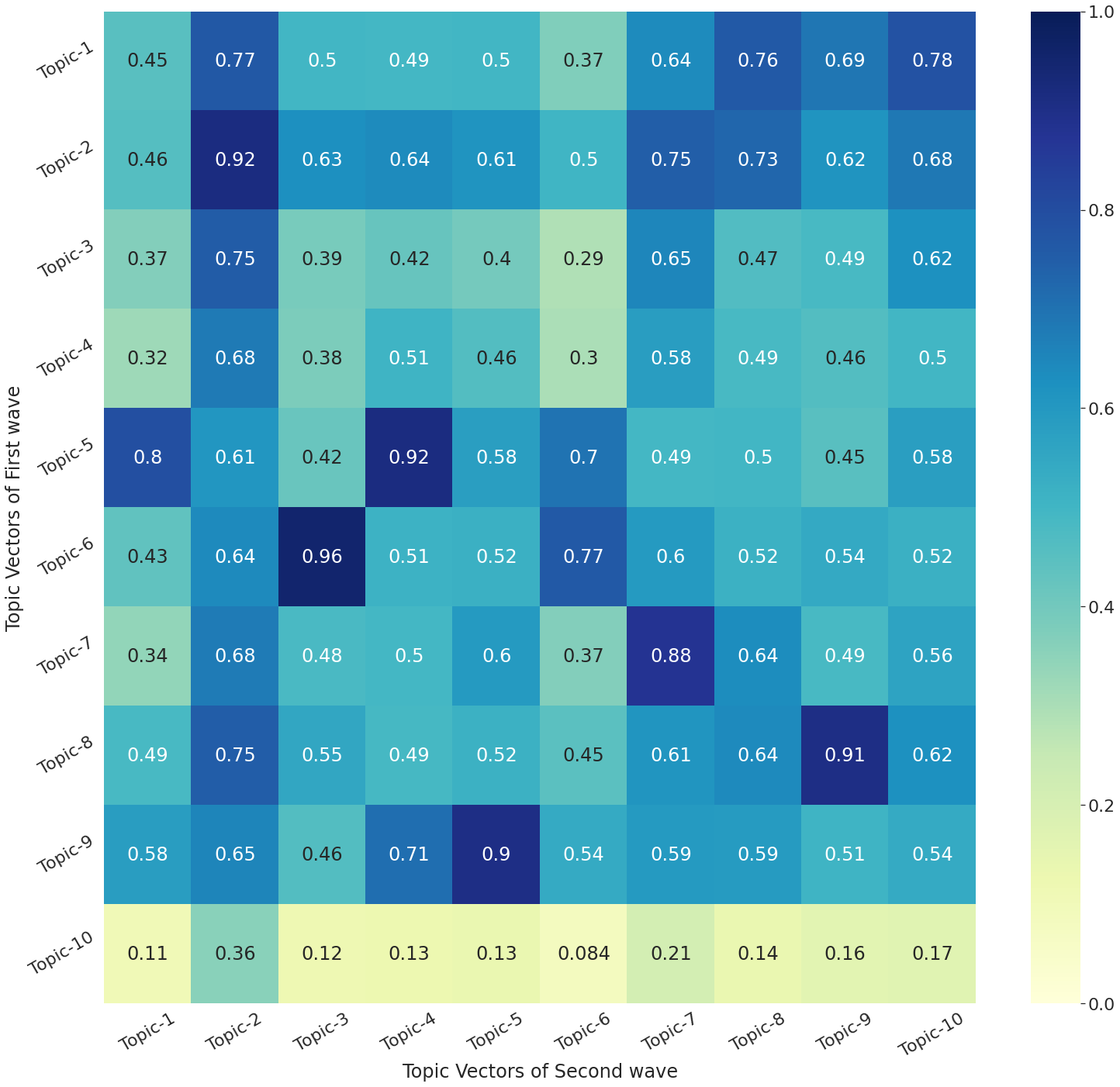}

\caption{Heatmap showing the similarity score  between different topics of First Wave  and Second Wave of COVID-19 generated using BERT-TM.}
\label{fig:heatmap1}
\end{figure}

We carry on further investigations using the BERT-TM which gave the best topic coherence results in the previous section. 
In order to evaluate the relationship between the first and the second wave of COVID-19, we use the topics obtained to find a similarity matrix  and present a heatmap (Figure \ref{fig:heatmap1}) to establish which topics from first wave are highly correlated to the topics from the second wave, and vice-versa. The similarity score was computed by cosine similarity. The heatmap represents the cosine similarity of the topic-vector obtained by the topic model. Therefore, in each  of the topics obtained from BERT-TM, we calculate its similarity with all the topics of the Upanishads and then find the topic with maximum similarity. There are various other measures of similarity score between two vectors; however, the cosine similarity is used widely in the literature \cite{salicchi2021pihkers,thongtan2019sentiment,gunawan2018implementation}. One of the major reasons for this is its interpretability.  Note that the value of cosine similarity between any two vectors lie between 0 and 1. A value closer to 1 represents perfect similarity and a value closer to 0 represent that they are completely dissimilar. The cosine similarity between any two vectors $U$ and $V$ is represented by Equation \ref{eq:cosine}. Since the topic vector contains contextual and thematic information about a topic, the similarity score gives us extent of closeness of the themes and topics of the Covid-19 waves.

\begin{ceqn}
\begin{align}
   \text{Sim(U, V)}=\cos(\theta )= \frac{\mathbf {U} \cdot \mathbf {V}}{\|\mathbf {U} \|\|\mathbf {V} \|}
   \label{eq:cosine}
\end{align}
\end{ceqn}

Next, we evaluate and discuss similar relationships both quantitatively using a mathematical formulation and also qualitatively by looking at the topics generated by our models as shown in Tables \ref{firstsecond}, and Figure \ref{fig:heatmap1}, the vertical axis of the heatmap shows the topics of the first wave while the horizontal axis of the heatmap represent the topics of the second wave.  We use the topics obtained to find a similarity matrix as shown in the heatmap and can observe from the Table \ref{firstsecond} that several topics in the first wave are similar to the topics of the second wave with more than 90\% similarity. 

 Table \ref{firstsecond} presents top 10 topics with 10 keyboards extracted and compares the first wave and second wave in terms of topic similarity score for  the COVID-19 pandemic in India. The highest similarity score is between Topic 6 (First Wave) and Topic 2 (Second Wave), followed by Topic 2 (First Wave) and Topic 3 (Second Wave) and Topic 5 (First Wave) and Topic 1 (Second Wave).  In comparison to the bigrams and trigrams in Figure \ref{bigramstrigrams}, we find a level of similarity in the top  topics extracted. We notice that the key words ['corona virus'], ['covid'], ['get well soon'], ['stay home safe'], ['narendramodi'], ['lockdown'] are common with topics extracted from  Table \ref{firstsecond} . The key words ['vaccination'], ['govt'], ['covid vaccines'], 'safe' are common with the topics extracted from the second wave.

Figure \ref{fig:heatmap1} heatmap shows that Topic 2 of the Second Wave is highly correlated to several topics of the First Wave (Topic 1, Topic 2, Topic 3 and Topic 8). Furthermore, we find that Topic 3 (Second Wave) is highly correlated to Topic 6 (First Wave), and Topic 4 (Second Wave) is highly correlated to Topic 5 (First Wave).  We review these topics with reference to Table 3 and find that majority of the topics deal with corona, vaccine shortage, government policies, hospitalisation, Prime Minister Narendra Modi, government officials, celebrities, corona updates, etc.

\subsubsection{Topics in media}
Next, we review the topics with emerging events and reports from the media during the first wave of COVID-19.
The first wave and second wave in India observed nationwide lockdown where the estimated economic cost of the Phase one lockdown of 21 days (March 25 to April 14, 2020) was estimated to be almost 98 billion dollars \cite{aprhit.2020}. The first wave lockdown in India was divided into four phases from March until the end of May 2020. In February 2021, India was hit by the largest COVID-19 wave. It is cited that people started becoming careless, not wearing masks and not following social distancing, around November- April. This wave caused a rapid surge in cases and deaths. Cases started to rise by March 2021, resulting in state-wide lock downs. In Maharashtra there were total phases phases of lock downs from April to June 2021. Due to large scale lock downs for a period of more than four months India observed both recession and unemployment, and majority of the masses suffered  (as shown in Table 3, Topic 1). 
 
 Indian Yoga Guru, Baba Ramdev  made controversial  comments about modern medicine and oxygen cylinders (May 2021). He particularly targeted allopathy medicines which was seen as a competition for Ayurveda medicine promoted by his company that serves as an alternative transitional Indian medicine   \cite{ramdev.2021}. There were calls in social media to arrest Baba Ramdev which is evident in Table 3 with Topic 2 keyword "arrestramdev".
The people showed widespread discontent towards government and policies during both the first and second waves, shown by Table 3 First Wave (Topic 8) and Second Wave (Topic 9) with the keywords such as "government", "corruption" and "bureaucrats".
The patients admitted to intensive care unit (ICU) during the second wave of the COVID-19 pandemic had significantly higher ICU and hospital mortality \cite{mortalityzirpe}, whilst both had high rate of hospitalisation. The keywords 'coronavirus',  'epidemic', "vaccine" and "vaccineshortage" were widely used throughout the pandemic period to describe COVID-19 as shown in Table 3  (Topics 3, 4,5 from First Wave) and (Topics 2 and 1 from Second Wave).  Table 3, Topic 2 keyword (Second Wave) "ripmilkhasingh" refers to the legendary Indian athlete Milkha Singh, also called 'flying Sikh' who was a four-time Asian Games gold medallist and 1958 Commonwealth Games champion  died of COVID-19 in June 2021 and was mourned throughout the country. 

India started to experience waves of recession with the coming of the first wave. Over 30\% of all industrial goods in India are transported via trains. Therefore, railway freight volumes become an important indicator of economic activity in the country. Given that many parts of India, including metro cities such as Mumbai and Delhi were under state government-imposed lockdowns, the daily average railway freight volumes in India dropped by 11\% month-on-month in April, according to Indian Railways data \cite{charted.2021}. One of the biggest impacts of the lockdowns in 2020 was a sharp rise in unemployment, especially in the unorganised sectors. In April 2020, unemployment in India spiked to 23 \%. However, as the country reopened, the job market started recovering and by February 2020, the unemployment rate fell to 6.9\%. In April 2021, the unemployment rate had gone up to 8.40\% \cite{unemploymentindia.2021}. These discussions are evident in Twitter as shown in Table 3, Topic 1 of First Wave and Topic 10 of Second Wave, with keywords about "lockdowns", "closed", "delayed", "collapse", "outbreaks" and "catastrophe". 

\begin{table*}[htbp!]
\scriptsize

\begin{adjustwidth}{-2.65in}{0in} 

\begin{tabular}{|l|l|l|l|r|}
\hline
Topics of First Wave(top 20 words) &
  \begin{tabular}[c]{@{}l@{}}First Wave\\ Topic ID\end{tabular} &
  Most Similar topics in Second Wave &
  \begin{tabular}[c]{@{}l@{}}Sec. Wave\\ Topic ID\end{tabular} &
  \multicolumn{1}{l|}{Score} \\ \hline
\begin{tabular}[c]{@{}l@{}}lockdown,locked,lockdowns,lockdownindia,lockdow,blocked,\\ lock,unlock,lack,prevent,daylockdown,closed,delayed,\\ unemployment,unlocked,badly,over,hence,rather,suffered\end{tabular} &
  topic - 1 &
  \begin{tabular}[c]{@{}l@{}}pandemic,pandemics,catastrophic,crisis,disaster,catastrophe,\\ epidemic,disasters,crises,recession,dengue,worrisome,panic,\\ rising,disastrous,outbreak,collapse,suffered,outbreaks,collapsed\end{tabular} &
  topic -10 &
  0.78 \\ \hline
\begin{tabular}[c]{@{}l@{}}kejriwal,amitabhbachchan,suspected,lakh,tested,suffered,\\ examination,bharat,haryana,hence,today,ahmedabad,\\jharkhand, amitabh,mukherjee,gandhi,recently,lakhs,\\chhattisgarh,coronaindia\end{tabular} &
  topic - 2 &
  \begin{tabular}[c]{@{}l@{}}covidisnotover,kejriwal,murshidabad,covidindia,allegedly,\\ breakingnews,ripmilkhasingh,covidsecondwave,covidvaccine,\\ shameonmpgovt,chiyaanvikram,covid,kharcha,kabir,suffered,\\ closely,lakh,seized,whereas,arrestramdev\end{tabular} &
  topic - 2 &
  0.91 \\ \hline
\begin{tabular}[c]{@{}l@{}}rather,hence,pathetic,thane,facepalming,toh,amitabhbachchan,\\ worry,than,suspected,which,instead,hdfc,meant,fm,shd,\\ means,suffered,wfh,amitabh\end{tabular} &
  topic - 3 &
  \begin{tabular}[c]{@{}l@{}}covidisnotover,kejriwal,murshidabad,covidindia,allegedly,\\ breakingnews,ripmilkhasingh,covidsecondwave,covidvaccine,\\ shameonmpgovt,chiyaanvikram,covid,kharcha,kabir,suffered,\\ closely,lakh,seized,whereas,arrestramdev\end{tabular} &
  topic - 2 &
  0.75 \\ \hline
\begin{tabular}[c]{@{}l@{}}coronaupdate,coronaupdates,corona,coronawarriors,\\coronaindia,coronalockdown,coronavaccine,coronapandemic,\\coronil,coron, coronavirus,king,colony,coro,skull,covaxin,\\covid,chaos,covidiots,covidwarriors\end{tabular} &
  topic - 4 &
  \begin{tabular}[c]{@{}l@{}}covidisnotover,kejriwal,murshidabad,covidindia,allegedly,\\ breakingnews,ripmilkhasingh,covidsecondwave,covidvaccine,\\ shameonmpgovt,chiyaanvikram,covid,kharcha,kabir,suffered,\\ closely,lakh,seized,whereas,arrestramdev\end{tabular} &
  topic - 2 &
  0.67 \\ \hline
\begin{tabular}[c]{@{}l@{}}coronavirus,chinesevirus,chinavirus,uhanvirus,vaccine,vaccines,\\ vaccination,virus,viruses,viruse,quarantined,viral,flu,epidemic,\\ viru,coronavaccine,infected,infect,infectious,quarantine\end{tabular} &
  topic - 5 &
  \begin{tabular}[c]{@{}l@{}}vaccineshortage,vaccineforall,vaccinemaitri,vaccinating,\\ vaccinefor,vaccinezehad,vaccinated,vaccineswork,vaccines,\\ vaccination,vaccine,vaccinate,vaccinat,vaccin,vaccinations,\\ vaccinequity,vaccinateindia,vacci,unvaccinated,getvaccinated\end{tabular} &
  topic - 1 &
  0.91 \\ \hline
\begin{tabular}[c]{@{}l@{}}indian,india,hindu,indians,hindustan,bharat,kejriwal,hindus,\\hindi, indi,crore,pakistan,gandhi,bangladesh,\\lockdownindia,caste, mukherjee,ghaziabad,crores,ahmedabad\end{tabular} &
  topic - 6 &
  \begin{tabular}[c]{@{}l@{}}indiahelp,indian,india,indianarmy,hindu,indians,hindutva,\\ hindustan,bharat,covidindia,kejriwal,vaccinateindia,hindus,\\ hindi,diwali,newindia,healthyindia,nehru,crore,pakistan\end{tabular} &
  topic - 3 &
  0.95 \\ \hline
\begin{tabular}[c]{@{}l@{}}appreciate,gratitude,blessed,grateful,appreciated,\\blessing, thankful,bless,blessings,helping,contribute,\\contributing,generous,honour,honoured,amitabhbachchan,\\honourable,thankyou,help, helps\end{tabular} &
  topic - 7 &
  \begin{tabular}[c]{@{}l@{}}shameonmpgovt,ripmilkhasingh,thanking,jharkhand,kejriwal,\\ blessed,kharcha,jammuandkashmir,grateful,diwali,bhadrak,\\ saintramrahimji,radheshyam,appreciate,blessing,murshidabad,\\ ganesh,lakh,gandhi,akshaykumar\end{tabular} &
  topic - 7 &
  0.87 \\ \hline
\begin{tabular}[c]{@{}l@{}}governments,government,govt,govts,gov,governance\\parliament,politicians,authorities,ministers,minister,\\politician,governor, officials,elected,administration,\\ruled,corruption,ministry,republic\end{tabular} &
  topic - 8 &
  \begin{tabular}[c]{@{}l@{}}governments,government,govt,govts,gov,governance,corruption,\\ bureaucrats,parliament,politicians,parliamentary,repeal,\\ farmersprotest,corrupt,federalism,politician,thepolitics,\\governor, administration,democracy\end{tabular} &
  topic - 9 &
  0.9 \\ \hline
\begin{tabular}[c]{@{}l@{}}hospitals,hospital,medical,patients,healthcare,clinical,nurse,\\ doctors,nurses,doctorsday,nursing,ambulance,medicine,\\ patient,doctor,clinic,cure,cured,dr,illness\end{tabular} &
  topic - 9 &
  \begin{tabular}[c]{@{}l@{}}hospitals,medical,hospital,hospitalised,hospitalized,healthcare,\\ hospitalization,hospitalisation,ambulance,patients, ambulances,\\ nurse,doctors,medicine,healthyindia,nurses,clinical,medicos,\\ dr,doctor\end{tabular} &
  topic - 5 &
  0.9 \\ \hline
\begin{tabular}[c]{@{}l@{}}havoc,sood,coz,hrs,tht,apne,kumar,zany,amit,ble,monday,\\ om,gtu,pic,uttarakhand,jamaat,kerala, kalyan,wuhan,setu\end{tabular} &
  topic - 10 &
  \begin{tabular}[c]{@{}l@{}}covidisnotover,kejriwal,murshidabad,covidindia,allegedly,\\ breakingnews,ripmilkhasingh,covidsecondwave,covidvaccine,\\ shameonmpgovt,chiyaanvikram,covid,kharcha,kabir,suffered,\\ closely,lakh,seized,whereas,arrestramdev\end{tabular} &
  topic - 2 &
  0.36 \\ \hline

\end{tabular}
\caption{Similarity score showing the inter-topic comparison between top 20 key words of First Wave and Second Wave in India.}
\label{firstsecond}

\end{adjustwidth}
\end{table*}

\subsection{Topic Modelling: Second vs Third wave}

\begin{figure}[htbp!]
\centering

\begin{adjustwidth}{-2.25in}{0in} 

\includegraphics[width=9cm]{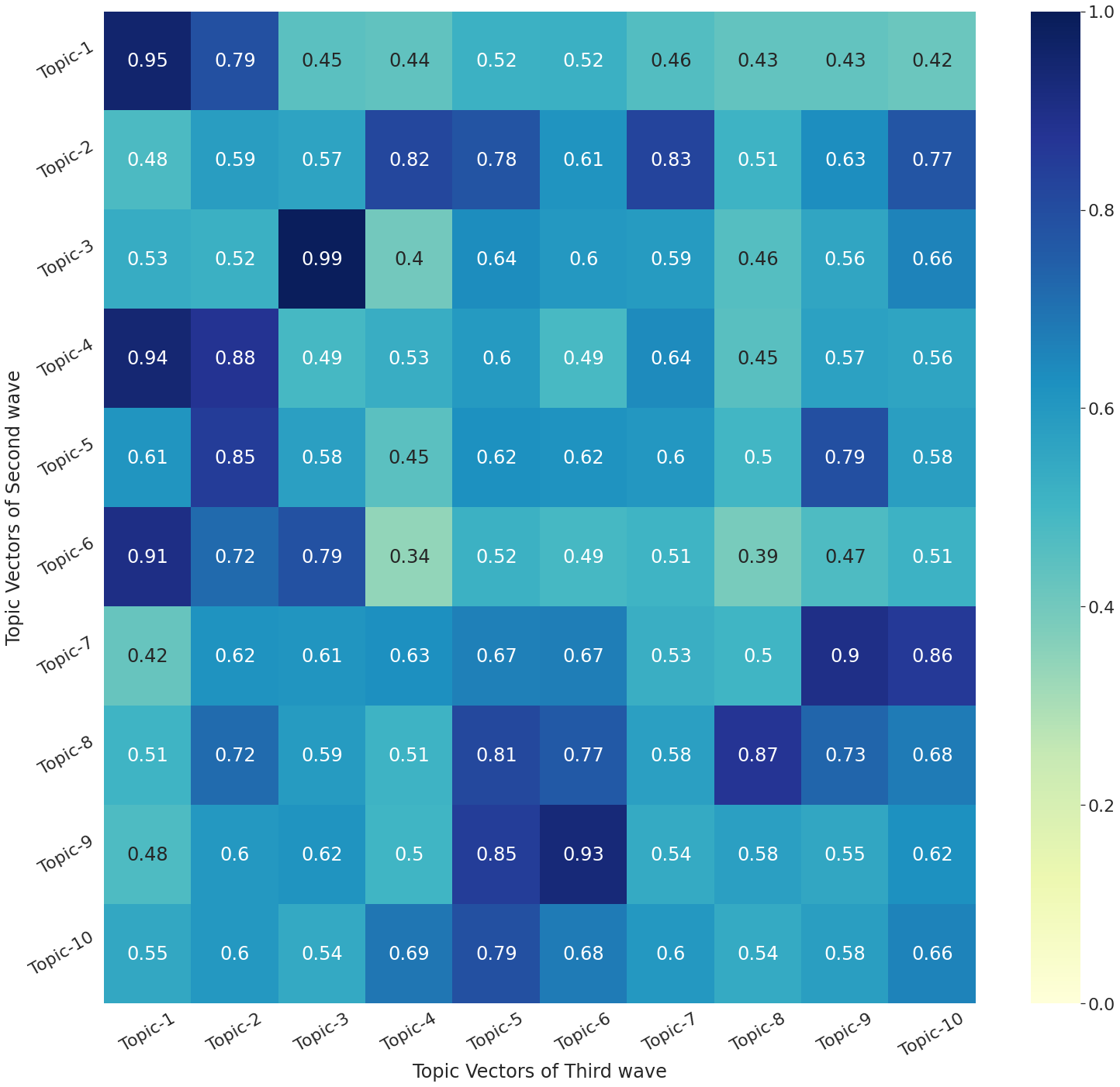}
\caption{Heatmap showing the similarity score  between different topics of Second Wave  and Third Wave of COVID-19 generated using BERT-TM.}
\label{fig:heatmap2}

\end{adjustwidth}
\end{figure}

We continue with our results from previous section that compared first and second waves. We can now compare the second and third waves of COVID-19 in India using the similarity matrix  presented as a heatmap (Figure \ref{fig:heatmap2}) where the vertical axis shows the topics of the Second Wave while the horizontal axis represents the topics of the Third wave. The heatmap shows that Topic 1 of the Third Wave is highly correlated with Topics 1,4 and 6 of the Second Wave. Further, coincidentally  Topic 3 of the second wave is highly correlated with Topic 3 of the third wave. Majority of the topics show similarity between vaccine-related issues, virus, celebrity news, universities, hospitalisation, prayers etc as shown in keywords for respective topics in Table \ref{secondthird}.

\subsubsection{Topics in media}

Next, we discuss how the topics relate to the media coverage for second wave of COVID-19 in India. The second and third wave in India had major overlapping topics over 'vaccination', 'vaccines', 'vaccinework' as shown in Table \ref{secondthird}, Topics 1,4, and 6 of the second wave and Topics 1 and 2 of the third wave.  After the second wave, the demand and intake of vaccines started increasing but soon enough due to India's drowsy vaccination programme a shortage was observed. India tried to increase the vaccine doses by banning its export of vaccines for a month \cite{covidlookslike.2021}. The central government asked the vaccine manufacturers to sell their doses at a lower price and allowed them to sell their vaccines to the private healthcare companies at a higher price to compensate. Unfortunately, the private healthcare system didn't get any incentive for the vaccines from the government. Moreover, the government limited the private hospital's profit margins. As a result, only large private healthcare companies were interested in the scheme, and  the state governments were constantly complaining that they aren't getting vaccines themselves (as shown in Table \ref{secondthird}: keywords in Topic 1 overlapping both in Second and Third Waves). 

Given the possibility of a third wave, in November 2021 the Indian Army increased its medical capacity across the country, while also helping the civilian administration in tackling the coronavirus (as shown in Table \ref{secondthird}, Topic 3 of Third Wave). The armed forces registered  about 200 COVID-19 cases every day, with the Army alone accounting for about 140 of them. But, significantly, most of these cases were mild and haven’t required hospitalisation \cite{armycovid.2021}. Board exams were cancelled \cite{examscancelled} as evident with keywords of Topic 8, both in Second and Third Waves of Table 4; "Students should not be forced to appear for exams in such a stressful situation," said the Indian Prime Minister, adding that all stakeholders need to show sensitivity for students. 

  The second wave began to witness signs of recession as evident by keywords, "recession", "unemployment" and "crisis" (Table 4, Topic 10 in Second Wave and Topic 5 in Third Wave). As per the official data released by the ministry of statistics and program implementation, the Indian economy contracted by 7.3 \% in the April-June quarter of this fiscal year \cite{recession.2021}. This was the worst decline ever observed since the ministry had started compiling GDP stats quarterly in 1996. India’s economy  shuttered during the lockdown period, except for some essential services and activities. As shops, eateries, factories, transport services, business establishments were shut, the lockdown had a devastating impact on slowing down the economy. The informal sectors of the economy have been worst hit by the global epidemic.

\begin{table*}[htbp!]
\scriptsize

\begin{adjustwidth}{-2.55in}{0in} 

\begin{tabular}{|l|l|l|l|r|}
\hline
Topics of Second Wave &
  Topic ID &
  Most Similar topics in Third Wave &
  Topic ID &
  \multicolumn{1}{l|}{Score} \\ \hline
\begin{tabular}[c]{@{}l@{}}vaccineshortage,vaccineforall,vaccineswork,vaccines\\ vaccinezehad,vaccinemaitri,vaccinefor,vaccinating,\\ vaccinations,vaccine,vaccination,vaccinate\\ 
vaccinated,vaccinat,vaccin,vaccinequity,vacci,\\ vaccinateindia,unvaccinated,getvaccinated\end{tabular} & topic-1 &
\begin{tabular}[c]{@{}l@{}}vaccinemandate,vaccines,vaccinating, \\ vaccination, vaccineswork, vaccine, vaccinate,\\ vaccinated, vaccinations, vaccinesuccess,vaccine,\\ vaccinequity, vaccinates, vacci, unvaccinated, getvaccinated,\\ fullyvaccinated, wuhanvirus, ebola, antiviral\end{tabular} &
topic - 1 &
0.95 \\ \hline
\begin{tabular}[c]{@{}l@{}}covidisnotover,kejriwal,murshidabad,covidindia, allegedly,\\ breakingnews,ripmilkhasingh,covidsecondwave,covidvaccine,\\ shameonmpgovt,chiyaanvikram,covid,kharcha,kabir,\\ suffered,closely,lakh,seized,whereas,arrestramdev\end{tabular} &
  topic-2 &
  \begin{tabular}[c]{@{}l@{}}newsupdate,latestnews,newstoday,newsupdates,updatenews,\\ middaynews,dailynews,breakingnews,news,indianews,noticias,\\ deaths,cases,newsoftheday,recently,worldcancerday,recent,\\ ommcomnews,fatalities,coronaviruses\end{tabular} & topic - 7 & 0.82 \\ \hline
\begin{tabular}[c]{@{}l@{}}indiahelp,indian,india,indianarmy,hindu,indians\\hindutva,hindustan, bharat,covidindia,kejriwal,\\vaccinateindia,hindus,hindi,diwali,newindia,\\ healthyindia,nehru,crore,pakistan\end{tabular} &
  topic-3 &
  \begin{tabular}[c]{@{}l@{}}indianews,indian,indiangovt,india,indianarmy,indias,hindu,\\ indianeconomy,indians,indiamap,indianrailways,indiablooms,\\ hindustani,hindutva,hindustan,bharat,covidindia,bharati,kejriwal,\\ hindi\end{tabular} &
  topic - 3 &
  0.99 \\ \hline
\begin{tabular}[c]{@{}l@{}}vaccineshortage,vaccineforall,vaccinemaitri \\vaccinating,vaccinefor,vaccinezehad,vaccinated, \\vaccineswork,vaccines,vaccination,vaccine,\\ vaccinate,vaccinat,vaccin,vaccinations,vaccinequity,\\vaccinateindia, vacci,unvaccinated,getvaccinated\end{tabular} &
  topic-4 &
  \begin{tabular}[c]{@{}l@{}}vaccinemandate,vaccines,vaccinating,vaccination,vaccineswork,\\ vaccine,vaccinate,vaccinated,vaccinations,vaccinesuccess,vaccin,\\ vaccinequity,vaccinates,vacci,unvaccinated,getvaccinated,\\ fullyvaccinated,wuhanvirus,ebola,antiviral\end{tabular} &
  topic - 1 &
  0.94 \\ \hline
\begin{tabular}[c]{@{}l@{}}hospitals,medical,hospital,hospitalised,\\hospitalized,healthcare, hospitalization,hospitalisation,\\ambulance,patients,ambulances, nurse,doctors,medicine,\\healthyindia,nurses,clinical,medicos,dr,\\ doctor\end{tabular} &
  topic-5 &
  \begin{tabular}[c]{@{}l@{}}vaccinemandate,vaccinated,vaccineswork,vaccines,vaccinating,\\ vaccine,vaccinate,vaccinesuccess,vaccin,vaccination,vaccinations,\\ vaccinequity,getvaccinated,fullyvaccinated,vacci,unvaccinated,\\ vaccinates,ebola,wuhanvirus,immunization\end{tabular} &
  topic - 2 &
  0.85 \\ \hline
\begin{tabular}[c]{@{}l@{}}vaccinateindia,vaccineshortage,vaccineforall, \\vaccines,vaccinezehad, vaccinations,vaccinating,vaccination,\\vaccinefor,vaccine,vaccineswork,vaccinate,vaccinated,\\vaccin,vaccinemaitri,vaccinat,vacci,vaccinequity,\\ unvaccinated,getvaccinated\end{tabular} & topic-6 &
\begin{tabular}[c]{@{}l@{}}vaccinemandate,vaccines,vaccinating,vaccination,vaccineswork,\\ vaccine,vaccinate,vaccinated,vaccinations,vaccinesuccess,vaccin,\\ vaccinequity,vaccinates,vacci,unvaccinated,getvaccinated,\\ fullyvaccinated,wuhanvirus,ebola,antiviral\end{tabular} &
  topic - 1 &
  0.9 \\ \hline
\begin{tabular}[c]{@{}l@{}}shameonmpgovt,ripmilkhasingh,thanking,jharkhand,\\kejriwal,blessed,kharcha,jammuandkashmir,\\grateful,diwali,bhadrak,saintramrahimji,\\ radheshyam,appreciate,blessing,murshidabad,ganesh,\\lakh,gandhi, akshaykumar\end{tabular} &
  topic-7 &
  \begin{tabular}[c]{@{}l@{}}gratitude,blessed,thanking,grateful,blessing,appreciate,\\ goodlucksakhi,bless,thankful,appreciated,wellbeing,condolences,\\ thanked,prayer,fortunately,thankfully,blessings,appreciating,\\ recover,prayed\end{tabular} &
  topic - 9 &
  0.9 \\ \hline
\begin{tabular}[c]{@{}l@{}}cancelboardexam, exams, sadly, examdate, exam, \\failed,  unfortunately,failing, caexams, suffered,\\ helpless, volunteers, unemployment, boardexams, didnt, \\volunteer, unemployed, students, delayed, sorry\end{tabular} &
  topic-8 &
  \begin{tabular}[c]{@{}l@{}}studentprotest, onlineexams, students, exams, aktuonlineexams,\\ cancelexams,exam,student,colleges,examinations,campus,studying,\\ cancelboardexam,classroom,universities,campuses,graduates,\\ semester,university,classrooms\end{tabular} &
  topic - 8 &
  0.87 \\ \hline
\begin{tabular}[c]{@{}l@{}}governments,government,govt,govts,gov,governance,corruption,\\ bureaucrats,parliament,politicians,parliamentary,repeal,\\ farmersprotest,corrupt,federalism,politician,thepolitics,\\ governor,administration,democracy\end{tabular} &
  topic-9 &
  \begin{tabular}[c]{@{}l@{}}parliament,parliamentary,governmental,governments,government,\\ politicians,corruption,govt,misgovernance,governance,\\ undemocratic,politician,govts,democracy,democratic,democrats,\\ elected,gov,bureaucrats,corrupt\end{tabular} &
  topic - 6 &
  0.93 \\ \hline
\begin{tabular}[c]{@{}l@{}}pandemic,pandemics,catastrophic,crisis,disaster,catastrophe,\\ epidemic,disasters,crises,recession,dengue,worrisome,panic,\\ rising,disastrous,outbreak,collapse,suffered,outbreaks,collapsed\end{tabular} &
  topic-10 &
  \begin{tabular}[c]{@{}l@{}}migrantcrisis,unemployment,overcoming,delayed,disrupted,\\ prevented,crisis,overdue,recession,migrants,boycottmodi,abruptly,\\ slowly,farmersprotest,badly,collapsed,preventing,prevent,\\ overwhelmed,sadly\end{tabular} &
  topic - 5 &
  0.79 \\ \hline

    \hline
\end{tabular}
\caption{Similarity score showing the inter-topic comparison between top 20 key words of Second Wave and Third Wave in India. }
\label{secondthird}

\end{adjustwidth}

\end{table*}

\subsection{Topic modelling: First vs Third wave}

\begin{figure}[htbp!]
\centering

\begin{adjustwidth}{-2.25in}{0in} 

\includegraphics[width=9cm]{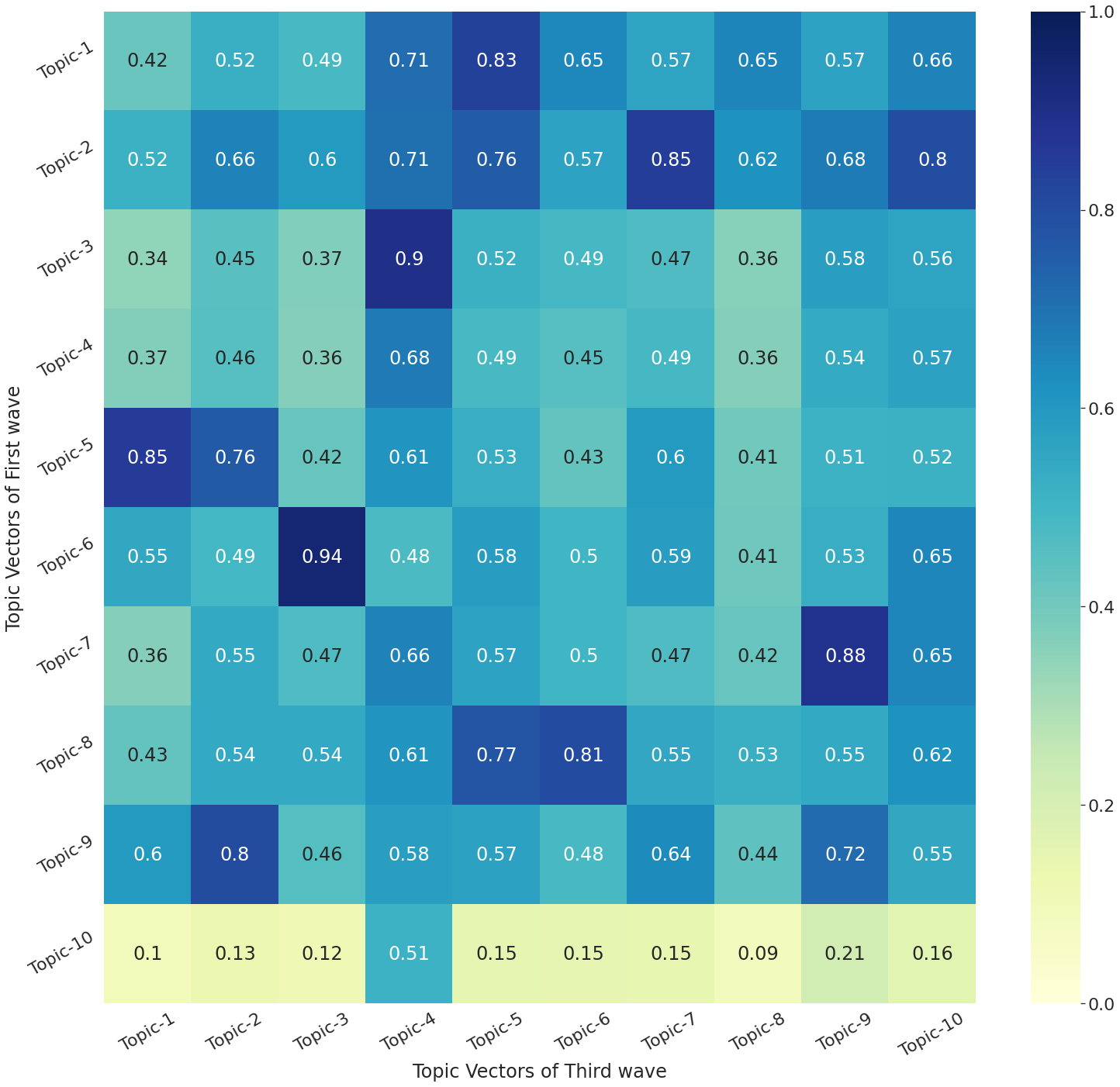} 
\caption{Heatmap showing the similarity score  between different topics of First Wave  and Third Wave of COVID-19 generated using BERT-TM.}

\label{fig:heatmap3}

\end{adjustwidth}
\end{figure}

Finally, we  compare the first and third waves of COVID-19 in India using the similarity matrix  presented as a  heatmap (Figure \ref{fig:heatmap3}).   This is to evaluate how the discussion about the COVID-19 pandemic evolved in Twitter and media, since the dynamics of the First Wave were very different when compared to the Third Wave. 
 
Figure \ref{fig:heatmap3} presents the heatmap that shows that Topic 1 of the Third  Wave is highly correlated to Topic 5 of the First Wave. The highest correlated is Topic 3 of the Third Wave with Topic 6 of the First wave and then there are several other combinations.  Table \ref{firstthird} presents topics of the Third Wave vs the First  Wave based on the similarity score as shown  in the previous section for other wave combinations. We notice that the highest correlated topic between the waves, as pointed out in Figure \ref{fig:heatmap3}, is given by keywords that have unique terms such as "indian, hindu, kejriwal, pakistan, gandhi, bangladesh, lockdownindia, caste" in First Wave vs "indianarmy, hindu, indianeconomy, indiamap, indianrailways, indiablooms, hindustani, hindutva" (Table \ref{firstthird}). This reveals the dynamic nature of COVID-19 and associated topics and the common issues (First and Third Waves) that prevailed in the pandemic that included nationalism, lockdowns, Hinduism which transformed to railways, Indian army and economy in the Third Wave.

\subsubsection{Topics in media}
 
It was reported  by the government that over 10 million ( one crore) \cite{workersreturn} inter-state migrant workers returned home on foot during the March to June 2020 during First Wave which had lockdowns  which activists called were an under-representation of scale of the crisis. The Indian scientists and research institutions were praised with two vaccines indigenously developed and manufactured and had been approved for emergency use in India with competitive level of efficacy when compared to western counterparts  \cite{kumar2021strategy,ahmed2022inactivated}. The world's largest vaccine drive was underway in the country and was moving forward at a rapid pace with more than 10 million doses administered already by the end of the second wave \cite{thirdwaveover2022}, although there were major challenges in terms of vaccination of rural areas and vaccine hesitancy \cite{chandani2021covid}.


\begin{table*}[htbp!]
\scriptsize

\begin{adjustwidth}{-2.25in}{0in} 

\begin{tabular}{|l|l|l|l|r|}
\hline
Topics of First Wave &
  Topic ID &
  Most Similar topics in Third Wave &
  Topic ID &
  \multicolumn{1}{l|}{Score} \\ \hline
\begin{tabular}[c]{@{}l@{}}lockdown,locked,lockdowns,lockdownindia,lockdow,blocked,\\ lock,unlock,lack,prevent,daylockdown,closed,delayed,\\ unemployment,unlocked,badly,over,hence,rather,suffered\end{tabular} &
  topic -1 &
  \begin{tabular}[c]{@{}l@{}}migrantcrisis,unemployment,overcoming,delayed,\\ disrupted,prevented,crisis,overdue,recession,migrants,\\ boycottmodi,abruptly,slowly,farmersprotest,badly,\\ collapsed,preventing,prevent,overwhelmed,sadly\end{tabular} &
  topic - 5 &
  0.83 \\ \hline
\begin{tabular}[c]{@{}l@{}}kejriwal,amitabhbachchan,suspected,lakh,tested,suffered,\\ examination,bharat,haryana,hence,today,ahmedabad,\\ jharkhand,amitabh,mukherjee,gandhi,recently,lakhs,\\ chhattisgarh,coronaindia\end{tabular} &
  topic -2 &
  \begin{tabular}[c]{@{}l@{}}newsupdate,latestnews,newstoday,newsupdates,\\ updatenews,middaynews,dailynews,breakingnews,\\ news,indianews,noticias,deaths,cases,newsoftheday,\\ recently,worldcancerday,recent,ommcomnews,fatalities,\\ coronaviruses\end{tabular} &
  topic - 7 &
  0.84 \\ \hline
\begin{tabular}[c]{@{}l@{}}rather,hence,pathetic,thane,facepalming,toh,amitabhbachchan,\\ worry,than,suspected,which,instead,hdfc,meant,fm,shd,means,\\ suffered,wfh,amitabh\end{tabular} &
  topic -3 &
  \begin{tabular}[c]{@{}l@{}}covidisnotover,covidguideline,covidpandemic,\\ covidvaccine,coviddeaths,cooch,covidvaccines,\\ covidguidelines,covid,covidhero,comorbid,covidtest,\\ covovax, covaxin, wbpc, covax, kkundrrasquad,\\ bymygov, sdm, covidpositive\end{tabular} &
  topic - 4 &
  0.89 \\ \hline
\begin{tabular}[c]{@{}l@{}}coronaupdate,coronaupdates,corona,coronawarriors,coronaindia,\\ coronalockdown,coronavaccine,coronapandemic,coronil,coron,\\ coronavirus,king,colony,coro,skull,covaxin,covid,chaos,covidiots,\\ covidwarriors\end{tabular} &
  topic -4 &
  \begin{tabular}[c]{@{}l@{}}covidisnotover,covidguideline,covidpandemic,\\ covidvaccine,coviddeaths,cooch,covidvaccines,\\ covidguidelines,covid,covidhero,comorbid,covidtest,\\ covovax,covaxin,wbpc,covax,kkundrrasquad,\\ bymygov,sdm,covidpositive\end{tabular} &
  topic - 4 &
  0.68 \\ \hline
\begin{tabular}[c]{@{}l@{}}coronavirus,chinesevirus,chinavirus,uhanvirus,vaccine,vaccines,\\ vaccination,virus,viruses,viruse,quarantined,viral,flu,epidemic,\\ viru,coronavaccine,infected,infect,infectious,quarantine\end{tabular} &
  topic -5 &
  \begin{tabular}[c]{@{}l@{}}vaccinemandate,vaccines,vaccinating,vaccination,\\ vaccineswork,vaccine,vaccinate,vaccinated,vaccinations,\\ vaccinesuccess,vaccin,vaccinequity,vaccinates,vacci,\\ unvaccinated,getvaccinated,fullyvaccinated,wuhanvirus,\\ ebola,antiviral\end{tabular} &
  topic - 1 &
  0.85 \\ \hline
\begin{tabular}[c]{@{}l@{}}indian,india,hindu,indians,hindustan,bharat,kejriwal,hindus,hindi,\\ indi,crore,pakistan,gandhi,bangladesh,lockdownindia,caste,\\ mukherjee,ghaziabad,crores,ahmedabad\end{tabular} &
  topic -6 &
  \begin{tabular}[c]{@{}l@{}}indianews,indian,indiangovt,india,indianarmy,indias,\\ hindu,indianeconomy,indians,indiamap,indianrailways,\\ indiablooms,hindustani,hindutva,hindustan,\\ bharat,covidindia,bharati,kejriwal,hindi\end{tabular} &
  topic - 3 &
  0.94 \\ \hline
\begin{tabular}[c]{@{}l@{}}appreciate,gratitude,blessed,grateful,appreciated,blessing,thankful,\\ bless,blessings,helping,contribute,contributing,generous,honour,\\ honoured,amitabhbachchan,honourable,thankyou,help,helps\end{tabular} &
  topic -7 &
  \begin{tabular}[c]{@{}l@{}}gratitude,blessed,thanking,grateful,blessing,appreciate,\\ goodlucksakhi,bless,thankful,appreciated,wellbeing,\\ condolences,thanked,prayer,fortunately,\\ thankfully,blessings,appreciating,recover,prayed\end{tabular} &
  topic - 9 &
  0.88 \\ \hline
\begin{tabular}[c]{@{}l@{}}governments,government,govt,govts,gov,governance,parliament,\\ politicians,authorities,ministers,minister,politician,governor,\\ officials,elected,administration,ruled,corruption,ministry,republic\end{tabular} &
  topic -8 &
  \begin{tabular}[c]{@{}l@{}}parliament,parliamentary,governmental,governments,\\ government,politicians,corruption,govt,misgovernance,\\ governance,undemocratic,politician,govts,democracy,\\ democratic,democrats,elected,gov,bureaucrats,corrupt\end{tabular} &
  topic - 6 &
  0.8 \\ \hline
\begin{tabular}[c]{@{}l@{}}hospitals,hospital,medical,patients,healthcare,clinical,nurse,\\ doctors,nurses,doctorsday,nursing,ambulance,medicine,patient,\\ doctor,clinic,cure,cured,dr,illness\end{tabular} &
  topic -9 &
  \begin{tabular}[c]{@{}l@{}}vaccinemandate,vaccinated,vaccineswork,vaccines,\\ vaccinating,vaccine,vaccinate,vaccinesuccess,vaccin,\\ vaccination,vaccinations,vaccinequity,getvaccinated,\\ fullyvaccinated,vacci,unvaccinated,vaccinates,ebola,\\ wuhanvirus,immunization\end{tabular} &
  topic - 2 &
  0.8 \\ \hline
\begin{tabular}[c]{@{}l@{}}havoc,sood,coz,hrs,tht,apne,kumar,zany,amit,ble,monday,om,gtu,\\ pic,uttarakhand,jamaat,kerala,kalyan,wuhan,setu\end{tabular} &
  topic -10 &
  \begin{tabular}[c]{@{}l@{}}covidisnotover,covidguideline,covidpandemic,\\ covidvaccine,coviddeaths,cooch,covidvaccines,\\ covidguidelines,covid,covidhero,comorbid,covidtest,\\ covovax,covaxin,wbpc,covax,kkundrrasquad,\\ bymygov,sdm,covidpositive\end{tabular} &
  topic - 4 &
  0.51 \\ \hline
\end{tabular}

\caption{Comparison between top 20 words of First Wave and Third Wave}
\label{firstthird}

\end{adjustwidth}
\end{table*}


\subsection{Further visualisation after hierarchical topic reduction}

\begin{figure}[htbp!]
\centering

\begin{adjustwidth}{-2.25in}{0in} 

\begin{subfigure}{0.42\textwidth}
    \includegraphics[width=\textwidth]{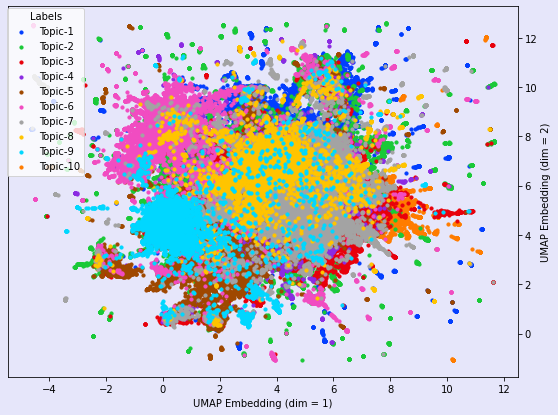}
    \caption{First wave} 
\end{subfigure}
\hfill
\begin{subfigure}{0.42\textwidth}
    \includegraphics[width=\textwidth]{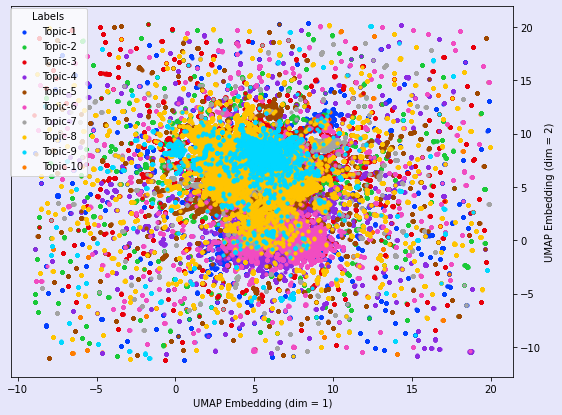}
    \caption{Second wave} 
\end{subfigure} 
\begin{subfigure}{0.42\textwidth}
    \includegraphics[width=\textwidth]{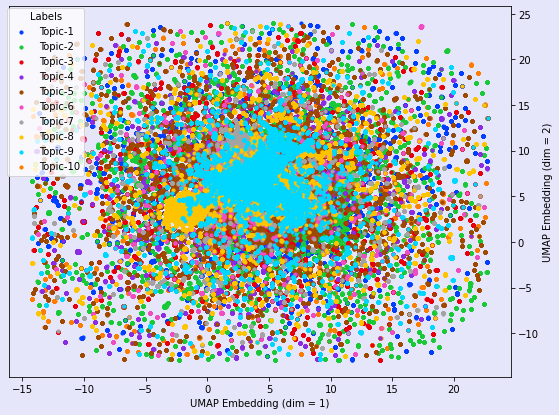}
    \caption{Third wave} 
\end{subfigure} 
\caption{Scatter plot of UMAP embedding of the first, second and third waves after hierarchical topic reduction.}
\label{fig:umap-waves}

\end{adjustwidth}
\end{figure}

 Figure \ref{fig:umap-waves} presents  scatter plot of first two UMAP embedding of the first, second and third COVID-19 waves after implementing hierarchical topic reduction. We reduced the number of topics using hierarchical topic reduction \cite{angelov2020top2vec}. Since the number of documents and words are different for the different corpus as seen, the number of topics obtained are different for different corpus. We reduced the number of topics to 10 in order to visualize the topic's semantic space clearly while plotting the semantic space for the different topics obtained by our framework. 
 
 In Figure \ref{fig:umap-waves}  Panel (a), we notice that Topic 9 dominates along with Topic 3 for the first wave. Tables 3 and 4 presents the major topics that are linked to Figure \ref{fig:umap-waves}. In the case of the second wave, Figure \ref{fig:umap-waves}  Panel (b) shows that Topic 9 largely dominates and finally, for the third wave, we find that Topic 9 dominates. It is unclear what exactly these topics are but if we look at the similarity score from Tables 3 and 4, we can infer that the topics that have large clusters may relate to the topics that have higher similar scores across the waves. 

\section{Discussion} 

India witnesses a number of events during these waves, which included several  regional elections, spanning 2020 - 2022 \cite{mahmood2022elections}, farmers protest\cite{behl2022india}, and roll-out of vaccines \cite{lancet2021india}. Our results show that the first, second and third wave observed a variety of overlapping as well as distinct topics. India suffered from forced lockdowns  and the closure of borders, and diplomatic relations with other countries also suffered. Although social media played a vital role in the pandemic, there was often alterations in dissemination of reports from the authorities, resulting to misinformation in social media \cite{cinelli2020covid} which had positive and negative impacts \cite{venegas2020positive}, which was not constrained to India.
 
Our results indicate that the major topics during the first wave features lockdowns, economic crisis, school closures, vaccines, government policies, the reaction of the people, death polls, donations, celebrities in India, doctors, hospitals, and religion. The second wave saw a rise in the number of topics related to vaccine such as Covaxin with the  the age group wise vaccine drive in India \cite{chakraborty2021current}. We found that vaccination, hospitalisation and governance were central to the topics from the  discussions \cite{jain2021differences}. We note that the second wave was the most severe in India due to the Delta variant \cite{bian2021impact} with higher rate of infections and death rate \cite{sarkar2021covid}. In the third wave, with the Omicron variant \cite{he2021sars}, the topics ranged from vaccination, governance, to economic recovery which marked the end of the pandemic as restrictions were eased and travel become normal with less restrictions. This was mainly because India had a high rate of vaccination by the third wave, which was less severe than first and second waves as the country was well prepared in medical supplies and management of hospitals \cite{ranjan2022omicron}. 

We note that a large effort was made in downloading using Twitter tool known as hydrator which normally has several restrictions, as large amounts of data cannot be downloaded at once. Hence our team had to manually download and check the data download process at regular intervals and the processes took more than six months of our time. However, we managed to publish the data (password protected to suit Twitter policy) in Kaggle \cite{janhavi2022}. Our data covers the COVID-19 pandemic from March 2020 i.e. from the emergence of COVID-19 capturing major Twitter active countries such as the USA, UK, Brazil, India, Japan, Indonesia, Australia, and Indonesia. In this paper, we restricted the study to India, however, in future works the study can extend to other countries. It would be interesting to compare the different countries at the different phases of the pandemic. 

In terms of limitations, we note that the data source considered was not taken daily but taken on three selected days of a week. Moreover, it is difficult to apply topic modelling methods on tweets since they are restricted by size and also includes everyday language expressions that rely on local regions and also influenced by regional languages in India. Although the data is sourced in English, we note that most of India has English as a second language and there are a number of regional languages in India. The 2011 Indian census recorded 31 regional Indian languages (such as Hindi, Bengali, Tamil, and Punjabi) which had at least one million speakers each \cite{chandras2020multilingualismin}, this gives a better picture of the language diversity in India which is a major challenge when it comes to translations \cite{kothari2014translating}.  Hence, there would be a bias in expression, with terms that are associated with regional languages according to the Tweet user background. Language translation for Indian languages has been of interest \cite{khan2019statistical} along with speech recognition for Indian languages \cite{singh2020asroil}.

 The topic coherence score  (NPMI)  is an approximate measure which can change for different types of documents and  we  need qualitative studies to further validate the topics. In our study, we validated selected topics extracted using media sources during the different waves of the pandemics. However, this is not a systematic approach and a major challenge of topic modelling method is validation of results. We need to develop methods using advanced methods. Recently, there has been much emphasis on ChatGPT(Chat Generative Pre-trained Transformer) \cite{chatGPT} - which is a NLP chatbot that provides answers to questions \cite{zhang2019dialogpt} which has a wide range of applications, that includes medicine and healthcare \cite{korngiebel2021considering}, with raise in concerns about academic plagiarism \cite{dehouche2021plagiarism}. Perhaps in future research, ChatGPT can be used as a mechanism to evaluate topics extracted from related studies, given that ChatGPT retrieves news media sources. However, we note that current version of ChatGPT does not provide comprehensive referencing, we do not know from where the the informed has been sourced. Thus, in future if ChatGPT or related models are improved, it can be useful for topic modelling.

\section{Conclusions} 

 In this paper, we presented a topic modeling framework for COVID-19 topic modelling in India via Twitter. We first compared BERT-based topic modelling with conventional approaches and found that BERT-based topic modeling performs better in terms of topic coherence. Hence we used it further to extract topics from the three major waves of India and student the correlation of major topics between the different waves and reported topics that were distinct for particular waves and also prominent throughout the pandemic.  We found a strong correlation of some of the the topics qualitatively to news media prevalent at the respective time period. Our topic modelling framework has been effective and provides a systematic methodology of understanding the major topics of discussion in social media that covers governance, vaccination, and management of COVID-19 pandemic with challenges and fear that includes the lockdowns and effect on the economy.

\section*{Data and Code}

Python-based open source code and data can be found here:
 \url{https://github.com/sydney-machine-learning/topicmodelling-COVID19-India}.

\section*{Author contributions statement}
 J. Lande  provided implementation and experimentation and further contributed in results visualisation and analysis. A. Pillay provided literature review and analysis of results. R. Chandra devised the project with the main conceptual ideas and experiments and contributed to overall writing,literature review and discussion of results.


\end{document}